\documentclass[11pt]{article}

\usepackage[final]{acl}

\usepackage{times}
\usepackage{latexsym}

\usepackage[T1]{fontenc}

\usepackage[utf8]{inputenc}

\usepackage{microtype}

\usepackage{inconsolata}

\usepackage{graphicx}
\usepackage{booktabs}
\usepackage{amsmath}
\usepackage{tcolorbox}

\title{Generating Effective CoT Traces for Mitigating Causal Hallucination}

\author{
Yiheng Zhao\textsuperscript{1,\textdagger},
Jun Yan\textsuperscript{1} \\
\textsuperscript{1}Concordia University, Montreal, Canada \\
\textdagger\ Corresponding author: \href{mailto:yiheng.zhao@mail.concordia.ca}{yiheng.zhao@mail.concordia.ca}
}

\begin{document}
\maketitle
\begin{abstract}
Although large language models (LLMs) excel in complex reasoning tasks, they suffer from severe causal hallucination in event causality identification (ECI), particularly in smaller models ($\leq$1.5B parameters). A promising approach to address this issue is to fine-tune them with Chain-of-Thought (CoT) traces. However, there is currently a lack of CoT trace dataset available for ECI. In this paper, we first investigate the essential criteria that effective CoT traces should possess to mitigate causal hallucination in smaller models. We then design a pipeline to generate CoT traces that meet these criteria. Moreover, since there is currently no metric for quantifying causal hallucination, we also introduce a new metric, the Causal Hallucination Rate (CHR), to quantify causal hallucination, guide the formulation of effective CoT trace criteria, and validate the effectiveness of our pipeline. Our experiments show that fine-tuning with the CoT traces generated by our pipeline not only substantially reduces causal hallucination in smaller LLMs but also improves mean accuracy. Moreover, the fine-tuned models exhibit strong cross-dataset and cross-difficulty generalization, as well as robustness under misleading intervention prompts.
\end{abstract}

\section{Introduction}
\begin{figure}[htpb]
    \centering
    \includegraphics[width=0.85\linewidth]{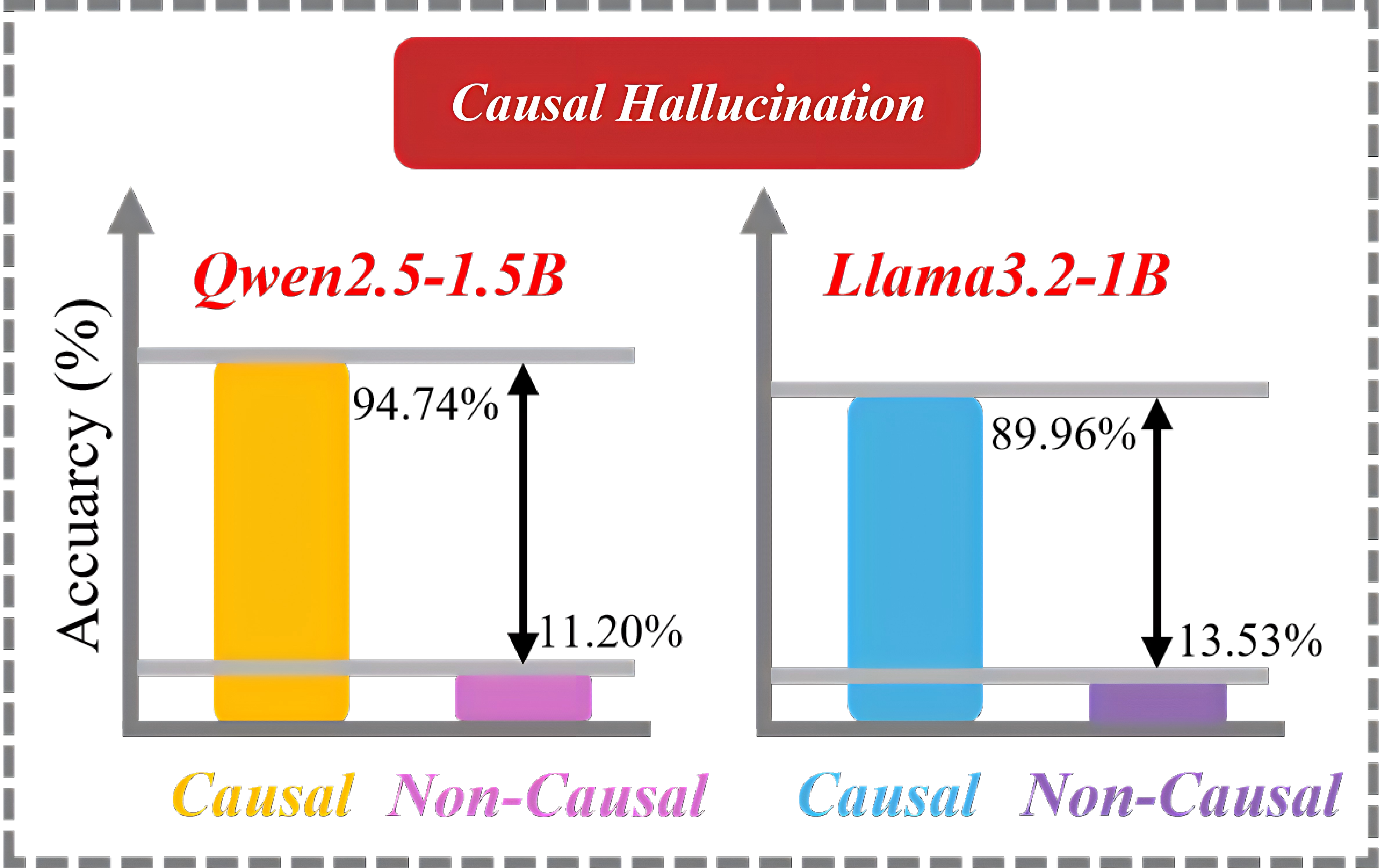}
    \caption{Illustration of causal hallucination in smaller models ($\leq$1.5B parameters) on EventStoryLine \cite{caselli-vossen-2017-event}, where a large accuracy gap exists between causal and non-causal event pairs. Causal event pairs denote pairs with a causal relationship, while non-causal pairs do not. Accuracy is computed separately for each category as the proportion of correctly predicted instances.}
    \label{fig:example-ch}
\end{figure}

Large language models (LLMs) have demonstrated impressive performance across a wide range of complex reasoning tasks, including mathematics, coding and deep research \cite{achiam2023gpt, yang2025qwen3, guo2025deepseek}. Despite these successes, recent studies reveal that they still suffer from severe causal hallucination \cite{gao2023chatgpt, cheng2025survey} in event causality identification (ECI), which aims to identify whether there is a causal relationship between two events in a text. Causal hallucination refers to a model's tendency to assume causal relationships between event pairs regardless of whether such relationships actually exist. We find this issue to be particularly pronounced in smaller models (Figure~\ref{fig:example-ch}); as smaller models are essential for real-world deployment, where efficiency matter, mitigating causal hallucination in these smaller models is critical.

A promising direction for addressing this issue is to fine-tune them using Chain-of-Thought (CoT) traces \cite{li2026instructiondataselectionanswer, li2026dataselectionmultiturndialogue}. However, current ECI datasets \cite{mirza2016catena, caselli-vossen-2017-event, wang2022maven, lai2022meci} contain only binary labels, which lack intermediate reasoning steps and are therefore inadequate for reducing causal hallucination. Moreover, existing LLM-based ECI studies primarily focus on prompt design and inference-time strategies, and there is currently no CoT trace dataset available for ECI.

In this paper, we first investigate the essential criteria that effective CoT traces should satisfy to mitigate causal hallucination in smaller models, focusing on three factors: perplexity, CoT trace length, and the distribution gap between CoT traces and the target models being fine-tuned. Prior works \cite{zhang2025best, li2025small, yang2024self} have proposed several criteria: (1) the one with the lowest perplexity should be preferred when multiple traces are available for a sample; (2) smaller models struggle to learn from long CoT traces; and (3) rewriting CoT traces with the target model to reduce the distribution gap helps to learn more effectively. However, we find that these criteria do not hold for ECI. Our analysis reveals three new findings: (1) perplexity is not a reliable selection criterion, as trace length has a stronger impact; (2) smaller models actually benefit more from long CoT traces; and (3) rewriting CoT traces is beneficial only when it does not increase perplexity. 

Based on these findings, we define that effective CoT traces for mitigating causal hallucination in smaller models should satisfy two criteria: (1) providing sufficiently long reasoning traces enriched with semantic explanations and intermediate steps; and (2) maintaining a small distribution gap and alignment with the target model. We then design the first data generation pipeline for ECI that generates fine-tuning data that meet these criteria, enabling smaller models to identify causal relationships more reliably. In addition, to properly quantify causal hallucination, we introduce the Causal Hallucination Rate (CHR), a new metric that quantifies causal hallucination, guides the formulation of effective CoT trace criteria, and validates the effectiveness of our pipeline. 

Experiments on EventStoryLine reveal that fine-tuning with the CoT traces generated by our pipeline yields significant reductions in causal hallucination while improving mean accuracy. Ablation analyses highlight the importance of rewriting CoT traces to match the target model's distribution. The fine-tuned models further exhibit strong cross-dataset generalization on Causal-TimeBank and MAVEN-ERE, as well as cross-difficulty generalization from sentence-level to document-level settings. Finally, robustness tests show that the models resist misleading intervention prompts, indicating that they have acquired more stable and reliable causal reasoning behavior. Together, these results demonstrate the effectiveness and broad applicability of our proposed pipeline for mitigating causal hallucination in smaller language models.

\section{Related Work}

\noindent \textbf{ECI with LLMs}\quad Existing LLM-based approaches primarily focus on inference-time strategies to enhance ECI performance of LLMs. Dr.ECI \cite{cai2025dr} constructs causal prompts grounded in causal-inference principles; MRBalance \cite{zou2025mrbalance} employs multi-agent debates with role-specific prompts to elicit complementary evidence; and MEFA \cite{zeng2025zero} decomposes ECI into six sub-tasks and designs dedicated prompts for each. In addition, Zhao et al. \cite{zhao2025can} investigate whether debate can improve the factuality and reasoning of LLMs on ECI.

\noindent \textbf{Building Effective CoT Traces}\quad Recent studies have explored how to construct effective fine-tuning data for improving LLM reasoning. However, these methods have been developed primarily in mathematical reasoning rather than for ECI. Zhang et al. \cite{zhang2025best} propose a perplexity-based selection strategy that chooses instruction–response pairs best aligned with the target model’s own distribution. Li et al. \cite{li2025small} show that small models cannot effectively learn from large teachers or long, complex reasoning traces, and thus introduce a mixed-distillation strategy for constructing more suitable fine-tuning data. Yang et al. \cite{yang2024self} further propose a rewriting-based self-distillation approach that generates model-aligned training data to narrow the distribution gap between downstream fine-tuning data and the target model.

\section{Methodology}
In this section, we first introduce the proposed causal hallucination metric. We then present the criteria for constructing effective CoT traces and describe our data generation pipeline used to produce CoT traces that meet the criteria.

\subsection{Causal Hallucination Rate (CHR)}
To mitigate causal hallucination, we must first quantify it. However, there is currently no metric for measuring causal hallucination in ECI. To address this gap, we propose CHR. This metric not only quantifies causal hallucination but also guides the formulation of effective CoT trace criteria and evaluates the effectiveness of our pipeline. The CHR is formulated as follows:
\begin{equation}
\text{CHR} = \text{Acc}_{\text{causal}} - \text{Acc}_{\text{non-causal}},
\label{eq:CHR}
\end{equation}
where $\text{Acc}_{\text{causal}}$ and $\text{Acc}_{\text{non-causal}}$ denote the model’s accuracy on causal and non-causal event pairs, respectively. A CHR value greater than 0 indicates causal hallucination, and smaller values reflect weaker hallucination. If CHR becomes negative, it means the model tends to overpredict non-causal relations, with larger deviations indicating stronger bias in that direction.

\subsection{Effective CoT Trace Criteria}
To investigate the criteria that effective CoT traces should satisfy to mitigate causal hallucination in smaller models, we first conduct a comprehensive analysis across three key factors: perplexity, CoT trace length, and the distribution gap between CoT traces and target models.

\noindent \textbf{Perplexity}\quad Prior work has shown that when multiple CoT traces exist for a sample, the one with the lowest perplexity should be selected. To verify whether this criterion holds in ECI, we conduct the following experiments. We first use two models (Qwen2.5-7B \cite{hui2024qwen2}, and Llama3.1-8B \cite{dubey2024llama}) to generate CoT traces and retain only those that produce correct answers. We choose models at the 7B–8B scale to avoid an excessively large distribution gap from the target models. We then fine-tune Qwen2.5-1.5B using (1) the traces from each model and (2) the lowest-perplexity traces selected across models. As shown in Table~\ref{ppl}, although the perplexity-based selection achieves the lowest perplexity, it does not yield the lowest CHR. Instead, longer traces (e.g., those generated by Llama) lead to substantially lower causal hallucination, indicating that length has a stronger influence than perplexity when multiple traces are available. To understand why longer traces are more effective, we examine their content and find that they contain richer semantic explanations, suggesting that semantic elaboration in CoT traces is the primary factor mitigating causal hallucination. Here, semantic explanations refer to additional explanations about the passage. To further validate this, we remove Qwen-generated traces from the perplexity-based selection, as they are shorter and contain fewer semantic explanations. As shown in Table~\ref{ppl}, the resulting CHR remains nearly unchanged, confirming that semantic explanation plays a crucial role. We report the following key takeaway.
\begin{tcolorbox}[
    colback=blue!10!white,
    colframe=gray!70,  
    boxrule=1.5pt        
]
\textbf{Takeaway for Perplexity:}
\textit{Perplexity is not a suitable selection criterion for ECI; length has a greater influence, with semantically richer CoT traces more effective at mitigating causal hallucination.}
\end{tcolorbox}
 
\begin{table}[t]
\centering
\small
\setlength{\tabcolsep}{10pt}
\begin{tabular}{l|ccc}
\toprule
Setting   & \textbf{PPL} & \textbf{Mean Token}  & \textbf{CHR}  \\
\midrule
Vanilla      &-& -  &84.55\\
Qwen   &2.77& 242  &60.30 \\
Llama  & 2.35& 317  &34.12\\
Perplexity  &2.28  & 302 &39.26 \\
LongOnly  & 2.28& - & 39.84 \\
\bottomrule
\end{tabular}
\caption{Results validating whether the perplexity-based selection strategy is effective. ``PPL" denotes perplexity, and ``Mean Token" represents the average token length of each CoT trace. ``Vanilla" denotes Qwen2.5-1.5B without CoT fine-tuning. ``Qwen" and ``Llama" use CoT traces generated by Qwen2.5-7B and Llama3.1-8B, respectively. ``Perplexity" selects the lowest-perplexity trace across models, while ``LongOnly" removes shorter Qwen traces from the perplexity-based selection set.}
\label{ppl}
\end{table}

\noindent \textbf{CoT Trace Length}\quad Previous study suggests that smaller models struggle to learn from longer CoT traces. However, our perplexity-based selection analysis contradicts this observation. Motivated by this observation, we further analyze the effect of trace length by fine-tuning Qwen2.5-1.5B using CoT traces of three different lengths. From Table~\ref{length}, we observe that CHR consistently decreases as CoT trace length increases, indicating that smaller models can effectively learn from longer reasoning sequences. In particular, traces containing rich semantic explanations and reasoning steps achieve the lowest CHR. This indicates that reasoning steps are essential for reducing causal hallucination, alongside semantic explanations. We summarize this observation as the following key takeaway.
\begin{tcolorbox}[colback=blue!10!white, colframe=gray!70, boxrule=1pt]
\textbf{Takeaway for CoT Trace Length:}
\textit{In ECI, smaller models benefit more from longer CoT traces; beyond richer semantic explanations, more reasoning steps are also crucial for mitigating causal hallucination.}
\end{tcolorbox}

\begin{table}[t]
\centering
\small
\setlength{\tabcolsep}{10pt}
\begin{tabular}{l|ccc}
\toprule
Setting  & \textbf{Mean Token} & \textbf{CHR} \\
\midrule
Vanilla     & -  &84.55\\
Qwen   & 242  & 59.79  \\
Llama  & 317  &34.68  \\
LlamaThinking  & 482 &30.60 \\
\bottomrule
\end{tabular}
\caption{Comparison of CoT traces with different lengths. ``Mean Token" represents the average token length of each CoT trace. ``Vanilla" denotes Qwen2.5-1.5B without CoT fine-tuning. ``Qwen'' uses concise traces from Qwen2.5-7B, ``Llama'' uses longer traces from Llama3.1-8B, and ``LlamaThinking'' uses even longer CoT traces generated by Llama3.1-8B, guided by few-shot examples produced by Qwen3-235B-A22B (Thinking) \cite{yang2025qwen3}.}
\label{length}
\end{table}

\noindent \textbf{Distribution Gap}\quad Although rewriting CoT traces using the target model has been shown to reduce the distribution gap and improve fine-tuning effectiveness, it remains unclear whether this strategy is beneficial for ECI, and at which CoT trace length rewriting yields the best performance. To investigate this, we evaluate the rewriting strategy, adopting the rewriting prompt template from prior work, across CoT traces of different lengths, as summarized in Table~\ref{dis-gap}. We observe that rewriting improves performance only for the shortest and longest CoT traces, while it increases CHR for medium-length traces. Further analysis shows that rewriting reduces perplexity for both short and long traces but increases it for medium-length traces, suggesting that the rewriting strategy is effective only when it does not increase perplexity. To verify this, we design a revised rewriting prompt, which successfully reduces perplexity and lowers CHR. Notably, rewriting the longest traces with richer semantic explanations and reasoning steps still yields the lowest CHR. 
\begin{tcolorbox}[colback=blue!10!white, colframe=gray!70, boxrule=1pt]
\textbf{Takeaway for Distribution Gap:}
\textit{Rewriting strategy to reduce the distribution gap mitigates causal hallucination only when perplexity does not increase; the best performance occurs when it is applied to CoT traces enriched with semantic explanations and reasoning steps.}
\end{tcolorbox}

\begin{table}[t]
\centering
\small
\setlength{\tabcolsep}{10pt}
\begin{tabular}{l|ccccccc}
\toprule
\textbf{Setting} & \textbf{PPL} & \textbf{CHR}  \\
\midrule
Vanilla      & -  &84.55\\
\midrule
Qwen   & 2.76   &59.79 \\
QwenR   & 2.75  &55.98  \\
\midrule
Llama  & 2.35  &34.68  \\
LlamaR  & 2.36   &42.81  \\
LlamaR (Our prompt)  & 2.34   &33.77 \\
\midrule
LlamaThinking  & 3.43  &30.60 \\
LlamaThinkingR  & 3.42   &24.42  \\
\bottomrule
\end{tabular}
\caption{Effect of applying rewriting on CoT traces of different lengths. 
``Qwen'', ``Llama'', and ``LlamaThinking'' follow the same definitions as in Table~\ref{length}. 
``QwenR'', ``LlamaR'', and ``LlamaThinkingR'' denote applying the prior study of rewriting prompt, while 
``LlamaR (our prompt)'' applies our revised rewriting prompt.}
\label{dis-gap}
\end{table} 

Based on the findings and takeaways discussed above, we formulate the following two essential criteria that effective CoT traces must meet to mitigate causal hallucination in smaller models.
\begin{tcolorbox}[
    colback=blue!10!white,
    colframe=gray!70,
    boxrule=1pt
]
\begin{itemize}
    \item \textbf{Criterion I:} \textit{CoT traces must contain sufficient semantic explanations and reasoning steps.}
    \item \textbf{Criterion II:} \textit{The distribution gap between CoT traces and the target models should be reduced without increasing perplexity.}
\end{itemize}
\end{tcolorbox}

In all analyses reported above, the training set is drawn from the first 16 topics of EventStoryLine, while the test set comprises the remaining four topics. The topics are processed in ascending order. To ensure fairness, the training samples are held identical across the scope of each factor analysis. In addition, the rewriting prompt template follows that of prior work, and our designed prompt is presented below.

\begin{tcolorbox}[
   colback=blue!10!white,
  colframe=gray!70,
  boxrule=1pt,
  arc=2pt,
  left=6pt,
  right=6pt,
  top=6pt,
  bottom=6pt
]
\textbf{Official instruction:}  
\textcolor{blue}{Below is an instruction that describes a task along with a reference answer.} Using the reference answer as a guide, write your own response.

\vspace{0.5em}
\textbf{Our designed instruction:}  
\textcolor{blue}{Rewrite the following response based on the given instruction and reference answer.} Using the reference answer as a guide, write your own response.

\vspace{0.7em}
\textbf{\#\#\# Instruction:}  
\texttt{\{instruction\}}

\vspace{0.5em}
\textbf{\#\#\# Reference:}  
\texttt{\{original CoT traces\}}

\vspace{0.5em}
\textbf{\#\#\# Response:}
\end{tcolorbox}

\subsection{CoT Trace Generation Pipeline}
\begin{figure}[htpb]
    \centering
    \includegraphics[width=0.9\linewidth]{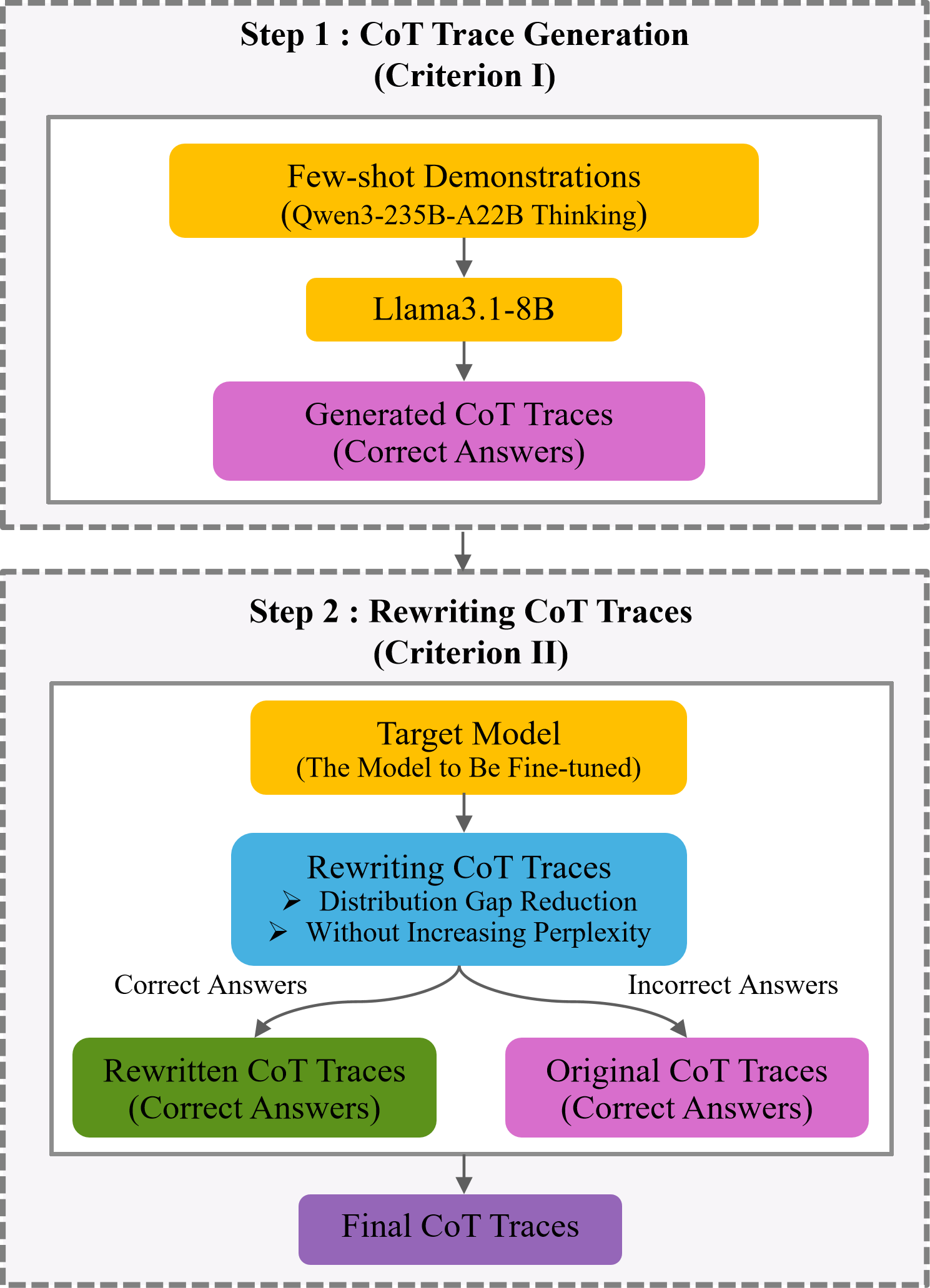}
    \caption{Overview of the proposed CoT trace generation pipeline.}
    \label{fig:pipeline}
\end{figure}

Based on the two criteria above, we design a two-step CoT trace generation pipeline tailored for mitigating causal hallucination in smaller models, as illustrated in Figure~\ref{fig:pipeline}, while maintaining low cost, as it primarily relies on Llama3.1-8B with few-shot demonstrations. In the first step, our goal is to produce CoT traces that meet \textbf{Criterion~I} by providing rich semantic explanations and reasoning steps. To achieve this, we employ Qwen3-235B-A22B (Thinking) to construct two few-shot demonstrations, one for a causal event pair and the other for a non-causal event pair. These demonstrations are then used to prompt Llama3.1-8B to generate the desired CoT traces, from which we retain only those that produce correct answers. In the second step, the goal is to reduce the distribution gap between the CoT traces generated in the first step and the target model. To achieve this, we rewrite the CoT traces using the target model itself and verify that the rewritten traces do not increase perplexity, thereby ensuring that \textbf{Criterion~II} is satisfied. Since our earlier analysis shows that the rewriting prompt from prior work does not increase perplexity for the targeted traces with rich semantic explanations and reasoning steps, we adopt this prompt in our pipeline. If a rewritten trace produces an incorrect answer, we retain the original trace from the first step instead of using the rewritten trace.

\section{Experiments}
In this section, we first introduce the datasets, implementation details, and evaluation metrics. We then present experimental results, including comparative analysis, ablation studies, and investigations of generalization and robustness.

\subsection{Datasets}
Our evaluation primarily utilizes three datasets: EventStoryLine, Causal-TimeBank, and MAVEN-ERE, employing different configurations for distinct experimental goals. 

For comparison, ablation, and robustness analyses, we consistently use sentence-level samples from the EventStoryLine dataset. Sentence-level means the input is strictly a single sentence. Across these three experiments, we use the same setup: we first sort the 20 topics in ascending order by topic ID, and subsequently perform 5-fold cross-validation for training and testing. It is important to note that, although the test set (the holdout set for each fold) remains complete, due to the generative capability limitations of models such as Llama 3.1-8B, the sample set used for model training includes only the subset for which the model successfully generated valid outputs. 

For cross-dataset generalization testing, we use sentence-level samples from the Causal-TimeBank and MAVEN-ERE datasets as our test sets. Furthermore, in the cross-difficulty generalization experiment, we adopt document-level samples from EventStoryLine as our test set. Document-level means the input is a full passage, and we focus exclusively on inter-sentence event pairs, i.e., events that occur in different sentences. For the causal event pairs, we use all available samples; for the non-causal event pairs, we randomly sample 10,000 pairs. More detailed dataset statistics are provided in the Appendix \ref{sec:data}.

\begin{tcolorbox}[
  colback=blue!10!white,
  colframe=gray!70,
  boxrule=0.8pt,
  arc=3pt,
  left=6pt,
  right=6pt,
  top=6pt,
  bottom=6pt
]
\textbf{Text:} An earthquake measuring at least magnitude-5.9 shook a sparsely populated area of southern Iran on Sunday, 
\textcolor{blue}{flattening} seven villages and \textcolor{blue}{killing} 10 people, officials said.

\vspace{0.4em}
\textbf{Intra-sentence event pairs:} 
(\textcolor{blue}{flattening}, \textcolor{blue}{killing})
\end{tcolorbox}

\begin{tcolorbox}[
  colback=blue!10!white,
  colframe=gray!70,
  boxrule=0.8pt,
  arc=3pt,
  left=6pt,
  right=6pt,
  top=6pt,
  bottom=6pt
]
\textbf{Text:} An earthquake measuring at least magnitude-5.9 shook a sparsely populated area of southern Iran on Sunday, 
\textcolor{blue}{flattening} seven villages and \textcolor{blue}{killing} 10 people, officials said. 
Tehran's seismological center said the \textcolor{blue}{quake} measured magnitude-5.9, but the U.S. Geological Survey in Golden, Colo., said it was a magnitude-6.1 temblor...

\vspace{0.4em}
\textbf{Inter-sentence event pairs:} 
(\textcolor{blue}{flattening}, \textcolor{blue}{quake}), 
(\textcolor{blue}{killing}, \textcolor{blue}{quake})
\end{tcolorbox}

Additionally, the two examples above illustrate the distinction between sentence-level and document-level ECI settings. In the sentence-level setting, the input is restricted to a single sentence, and the model evaluates only intra-sentence event pairs in which both events occur within the same sentence. In contrast, the document-level setting takes a passage as input and focuses on inter-sentence event pairs, where the two events appear in different sentences.

\subsection{Implementation Details and Metrics}

\noindent \textbf{Implementation Details}\quad All of our training is based on the SFTTrainer implementation within the Transformer Reinforcement Learning (TRL) framework. The specific training hyperparameters are configured as follows: the per-device batch size is 1, the gradient accumulation steps are 8, and the total training is conducted for 1 epoch. We set the learning rate to $2e-4$ and use a cosine-annealed scheduler. To enhance training efficiency and conserve computational resources, we adopt the Low-Rank Adaptation (LoRA) fine-tuning technique. The key parameters for the LoRA configuration are: the LoRA rank is 8, the scaling factor is 16, and the dropout rate is 0.05. More comprehensive training and implementation details are presented in the Appendix \ref{sec:training}. Moreover, in all experiments, we fix the decoding temperature to 0. This deterministic setting eliminates sampling randomness and ensures that all inference results are fully reproducible across runs. Moreover, in this paper, we fix the decoding temperature to 0 for all experiments.

\noindent \textbf{Metrics}\quad We use our proposed $\text{CHR}$ to evaluate causal hallucination. Additionally, we use mean accuracy ($\text{mAcc}$) to assess the models' overall performance. The $\text{mAcc}$ is defined as the average of the accuracy on causal event pairs and the accuracy on non-causal event pairs. 

\subsection{Results}

\noindent \textbf{Main Results}\quad To verify whether our proposed CoT trace generation pipeline can effectively reduce the causal hallucination in smaller language models, we test the effectiveness of our pipeline on two models: Qwen2.5-1.5B and Llama3.2-1B. Table $\ref{com-llms}$ presents the comparative results on EventStoryLine, contrasting the model fine-tuned on CoT trace data generated by our pipeline against various LLMs and existing baseline methods. For a fairer comparison with existing approaches, we re-implement these methods by replacing their original base LLMs with Qwen2.5-1.5B, which we use in our experiments, thereby allowing us to assess whether these methods can effectively reduce causal hallucination in smaller models.

\noindent From the Table~\ref{com-llms}, we can observe that existing methods such as Dr.ECI, MuTQA, and MRBalance do not reduce causal hallucination of Qwen2.5-1.5B. In particular, Dr. ECI with Qwen2.5-1.5B yields a CHR of 100.00\%, indicating that the method predicts a causal relation for every sample and increases causal hallucination. We also observe that prompt-based strategies, including zero-shot CoT prompting and ICL prompting, as well as fine-tuning with binary labels, fail to substantially mitigate hallucination. In contrast, fine-tuning the models on the CoT trace data generated by our pipeline yields a substantial reduction in hallucination for both models. Qwen2.5-1.5B achieves a CHR of 6.26\% and Llama3.2-1B reaches 9.14\%, corresponding to absolute reductions of 77.28\% and 67.29\% compared with the original models. After applying our method, the hallucination degree also becomes lower than that of much larger LLMs, including GPT-4, Llama3.1-8B and Qwen3-30B-A3B. Moreover, our pipeline not only suppresses causal hallucination but also improves overall reasoning accuracy, with mAcc rising from 52.97\% to 66.00\% for Qwen2.5-1.5B and from 55.58\% to 63.44\% for Llama3.2-1B. 
Further comparisons between model outputs before and after fine-tuning can be found in Appendix \ref{sec:vis}. To provide a more comprehensive evaluation, we also report additional metrics, including FPR, TNR, and MCC, with results presented in Appendix \ref{sec:add_evaluation}. 

\begin{table}[t]
\centering
\small
\setlength{\tabcolsep}{10pt}
\begin{tabular}{l|ccc}
\toprule
\textbf{Methods}    & \textbf{CHR}  & \textbf{mAcc}  \\
\midrule
GPT3.5-turbo     &43.43 & 57.36  \\
GPT4     &53.30 & 51.40  \\
Llama3.1-8B     &60.59    & 58.97 \\
Qwen3-30B-A3B     &60.27 & 58.76  \\
\midrule
Dr.ECI     &100.00 &50.00  \\
MuTQA      &46.03 &28.82  \\
MRBalance &82.45 &59.10  \\
\midrule
Qwen2.5-1.5B (Vanilla)    &83.54  &52.97  \\
Qwen2.5-1.5B (CoT)    &69.77  &51.48 \\
Qwen2.5-1.5B (ICL)    &69.57  &55.32  \\
Qwen2.5-1.5B (Binary)    & 66.67 &56.74  \\
Qwen2.5-1.5B (Our)    &\textbf{6.26} &\textbf{66.00}  \\
\midrule
Llama3.2-1B (Vanilla)    & 76.43 &55.58  \\
Llama3.2-1B (CoT)    &60.43  & 50.74 \\
Llama3.2-1B (ICL)    &58.42  &46.43  \\
Llama3.2-1B (Binary)    &56.55  &55.68  \\
Llama3.2-1B (Our)    & \textbf{9.14}&\textbf{63.44}  \\
\bottomrule
\end{tabular}
\caption{Comparison results on EventStoryLine. ``Vanilla'' refers to the original model; ``CoT'' denotes zero-shot chain-of-thought prompting; ``ICL'' denotes in-context learning with the same exemplars as those used to elicit long CoT traces from Llama3.1-8B in our pipeline; ``Binary'' refers to fine-tuning with binary labels; and ``Our'' indicates fine-tuning using CoT trace data generated by our proposed pipeline.}
\label{com-llms} 
\end{table}

\noindent \textbf{Ablation Results}\quad We further conduct an ablation study to examine the contribution of the rewriting step in our pipeline. We compare models fine-tuned on the original CoT traces, which contain rich semantic explanations and reasoning steps directly generated by Llama3.1-8B, against models fine-tuned on rewritten CoT traces produced by the target model itself. As shown in Table~\ref{ab-llms}, removing the rewriting step substantially weakens the ability to reduce causal hallucination. For Qwen2.5-1.5B, CHR decreases from 83.54\% to 30.39\% without rewriting, but further to 6.26\% with rewriting. A similar pattern is observed for Llama3.2-1B, where CHR decreases from 76.43\% to 17.13\% without rewriting, and to 9.14\% with rewriting. Moreover, rewriting yields notable improvements in mAcc for both models. Importantly, as shown in Table~\ref{ppl-results}, the rewriting step does not increase perplexity. These findings demonstrate that reducing the data distribution gap between the CoT traces and the target model, without increasing perplexity, is crucial for effectively suppressing causal hallucination.

\begin{table}[t]
\centering
\small
\setlength{\tabcolsep}{10pt}
\begin{tabular}{l|ccc}
\toprule
\textbf{Methods}    & \textbf{CHR}  & \textbf{mAcc}  \\
\midrule
Qwen2.5-1.5B (Vanilla)    &83.54  &52.97  \\
Qwen2.5-1.5B (w/o rewriting)    &23.39 &56.51  \\
Qwen2.5-1.5B (w rewriting)    &\textbf{6.26} &\textbf{66.00}  \\
\midrule
Llama3.2-1B (Vanilla)    & 76.43 &55.58  \\
Llama3.2-1B (w/o rewriting)    &17.13 &55.51 \\
Llama3.2-1B (w rewriting)    &\textbf{ 9.14}&\textbf{63.44}  \\
\bottomrule
\end{tabular}
\caption{Ablation results on EventStoryLine. ``Vanilla'' refers to the original model; ``w/o rewriting'' denotes fine-tuning without rewriting the initial CoT traces that contain rich semantic explanations and reasoning steps.  ``w rewriting'' denotes fine-tuning after rewriting these CoT traces using the target model.}
\label{ab-llms}
\end{table}

\begin{table}[t]
\centering
\small
\setlength{\tabcolsep}{10pt}
\begin{tabular}{l|cc}
\toprule
\textbf{Methods}    & \textbf{PPL}   \\
\midrule
Qwen2.5-1.5B (w/o rewriting)    &3.23   \\
Qwen2.5-1.5B (w rewriting)    &\textbf{3.18}   \\
\midrule
Llama3.2-1B (w/o rewriting)    &3.99 \\
Llama3.2-1B (w rewriting)    &\textbf{3.65}  \\
\bottomrule
\end{tabular}
\caption{Perplexity comparison with and without CoT trace rewriting.}
\label{ppl-results}
\end{table}

\subsection{Generalization}
\begin{table}[t]
\centering
\small
\setlength{\tabcolsep}{10pt}
\begin{tabular}{l|ccc}
\toprule
\textbf{Methods}    & \textbf{CHR}  & \textbf{mAcc}  \\
\midrule
\multicolumn{3}{c}{\textit{Causal-TimeBank}} \\\midrule
Qwen2.5-1.5B (Vanilla)    &84.55  & 51.53  \\
Qwen2.5-1.5B (Our)    &\textbf{ 11.37} &\textbf{ 66.79}  \\ \midrule
Llama3.2-1B (Vanilla)    &67.79  &45.85   \\
Llama3.2-1B (Our)    &\textbf{ 4.26} &\textbf{ 61.51} \\
\midrule
\multicolumn{3}{c}{\textit{MAVEN-ERE}} \\\midrule
Qwen2.5-1.5B (Vanilla)    &84.69  &52.50  \\
Qwen2.5-1.5B (Our)    &\textbf{ 11.13} &\textbf{ 64.92} \\ \midrule
Llama3.2-1B (Vanilla)    &73.82 & 50.79 \\
Llama3.2-1B (Our)    &\textbf{ 7.84}  &\textbf{ 62.86}  \\
\bottomrule
\end{tabular}
\caption{Cross-dataset results on Causal-TimeBank and MAVEN-ERE.  ``Vanilla'' refers to the original model without any fine-tuning. 
 ``Our'' denotes the model fine-tuned using the CoT trace data from the first fold of the five-fold cross-validation on EventStoryLine.}
\label{dataset-llms}
\end{table}

\noindent \textbf{Cross-dataset}\quad To evaluate the generalizability of the models fine-tuned, we further conduct cross-dataset experiments by applying the models fine-tuned on EventStoryLine to two external benchmarks: Causal-TimeBank and MAVEN-ERE. As shown in Table~\ref{dataset-llms}, the models fine-tuned achieve substantial reductions in causal hallucination alongside notable gains in predictive accuracy across both datasets. For Causal-TimeBank, the CHR of Qwen2.5-1.5B drops significantly from 84.55\% to 11.37\%, and its mAcc increases from 51.53\% to 66.79\%. Similarly, the CHR of Llama3.2-1B decreases from 67.79\% to 4.26\%, while mAcc improves from 45.85\% to 61.51\%. For MAVEN-ERE, Qwen2.5-1.5B reduces CHR from 84.69\% to 11.13\% and raises mAcc from 52.50\% to 64.92\%. Llama3.2-1B shows consistent improvements, with CHR dropping from 73.82\% to 7.84\% and mAcc rising from 50.79\% to 62.86\%. These results demonstrate that using our pipeline can generalize effectively to unseen corpora.

\begin{table}[t]
\centering
\small
\setlength{\tabcolsep}{10pt}
\begin{tabular}{l|ccc}
\toprule
\textbf{Methods}    & \textbf{CHR}  & \textbf{mAcc}  \\
\midrule
Qwen2.5-1.5B (Vanilla)    &54.52    &59.76  \\
Qwen2.5-1.5B (Our)    &\textbf{1.41}  &\textbf{69.85}  \\ \midrule
Llama3.2-1B (Vanilla)    &93.94 &51.62  \\
Llama3.2-1B (Our)    &\textbf{4.51}  &\textbf{59.45}  \\
\bottomrule
\end{tabular}
\caption{Document-level results on inter-sentence event pairs of EventStoryLine. ``Vanilla'' refers to the original model without any fine-tuning. ``Our'' denotes the model fine-tuned using the CoT trace data from the first fold of the five-fold cross-validation on EventStoryLine.}
\label{diff-llms}
\end{table}

\noindent \textbf{Cross-difficulty}\quad To further evaluate whether the models fine-tuned using the CoT trace data generated by our pipeline can generalize to more challenging scenarios, we additionally test them on the document-level ECI, where the input is a full passage, and the target event pairs occur across different sentences. This setting is substantially more difficult than the training-distribution sentence-level task due to longer contexts and more complex discourse structures. Using the models fine-tuned only on the first fold of EventStoryLine, we observe strong generalization improvements on this harder setting, as shown in Table~\ref{diff-llms}. For Qwen2.5-1.5B, CHR decreases significantly from 54.52\% to 1.41\%, while mAcc increases from 59.76\% to 69.85\%. For Llama3.2-1B, CHR decreases from 93.94\% to 4.51\%, while mAcc increases from 51.62\% to 59.45\%. These results demonstrate that the proposed pipeline not only suppresses causal hallucination within the training distribution but also generalizes effectively to substantially more challenging document-level ECI tasks.

\subsection{Robustness}

To assess whether these models fine-tuned, truly learn causal relationships in ECI, we also conduct a robustness evaluation. After training, we inject incorrect intervention into the prompt to mislead these models: for each causal event pair, we tell them that no causal relationship exists, and for each non-causal pair, we tell them that a causal relationship does exist. As shown in Table~\ref{ro-llms}, the mAcc of both models remains largely unchanged compared with the non-intervention setting, indicating that the fine-tuned models are robust and do not simply follow misleading instructions. The slight decrease in CHR arises because incorrect intervention prompts disrupt correct predictions on causal pairs and flip some incorrect predictions on non-causal pairs, leading to a small net shift in the metric. The specific non-intervention and incorrect-intervention prompts used for causal and non-causal event pairs are shown below.

\begin{tcolorbox}[
  colback=blue!10!white,
  colframe=gray!70,
  boxrule=0.8pt,
  arc=3pt,
  left=6pt,
  right=6pt,
  top=6pt,
  bottom=6pt
]

\vspace{0.5em}
\textbf{Instruction (Non-intervention for all event pairs):}  
Is there a causal relationship between \texttt{<event A>} and \texttt{<event B>}?

\vspace{0.5em}
\textbf{Instruction (Incorrect intervention for causal event pairs):}  
Is there a causal relationship...?  
\textit{You may refer to the provided information that there \textcolor{blue}{is no} causal relationship...}

\vspace{0.5em}
\textbf{Instruction (Incorrect intervention for non-causal event pairs):}  
Is there a causal relationship...?  
\textit{You may refer to the provided information that there \textcolor{blue}{is} a causal relationship...}
\end{tcolorbox}

\begin{table}[t]
\centering
\small
\setlength{\tabcolsep}{10pt}
\begin{tabular}{l|ccc}
\toprule
\textbf{Methods}    & \textbf{CHR}  & \textbf{mAcc}  \\
\midrule
Qwen2.5-1.5B (Vanilla)    &83.54  &52.97  \\
Qwen2.5-1.5B (Our)    &6.26 &66.00  \\
Qwen2.5-1.5B (Interv.)    &2.26 &63.08  \\
\midrule
Llama3.2-1B (Vanilla)    & 76.43 &55.58  \\
Llama3.2-1B (Our)    &9.14&63.44 \\
Llama3.2-1B (Interv.)    &4.34 &60.34 \\
\bottomrule
\end{tabular}
\caption{Robustness results on EventStoryLine. ``Vanilla'' refers to the original model without fine-tuning. ``Our'' denotes the model fine-tuned with our pipeline and evaluated without any intervention prompts. ``Interv.'' represents the same fine-tuned model evaluated under incorrect intervention prompts during inference.}
\label{ro-llms}
\end{table}

\section{Conclusions}

In this paper, we investigate how to mitigate causal hallucination in smaller LLMs for ECI. We first introduce the CHR, a simple yet informative metric that quantifies causal hallucination and guides both the analysis and evaluation of our methods. Through a systematic study of perplexity, CoT trace length, and distribution gap, we identify two key criteria for constructing effective CoT traces. Based on these insights, we develop a CoT trace generation pipeline that produces effective fine-tuning traces for smaller models, thereby substantially reducing their causal hallucination. Experiments on EventStoryLine show that fine-tuning with the CoT traces generated by our pipeline substantially reduces causal hallucination in smaller LLMs while also improving mean accuracy. The fine-tuned models further exhibit strong cross-dataset generalization to Causal-TimeBank and MAVEN-ERE, cross-difficulty generalization from sentence-level to document-level ECI, and robustness to misleading intervention prompts.

\section*{Limitations}
Despite the effectiveness of our approach, several limitations remain. First, our pipeline is primarily designed for smaller models ($\leq$1.5B parameters). While we demonstrate substantial gains in this regime, larger models are also known to exhibit causal hallucination, and extending our pipeline to mitigate hallucination in large-scale models remains an important direction for future work. Second, our evaluation of generalization focuses on cross-dataset and cross-difficulty settings within ECI. However, causal hallucination is also prevalent in other causal reasoning tasks, such as causal discovery, which are not explored in this work. Finally, although our robustness analysis shows that the fine-tuned models are resistant to deliberately misleading intervention prompts, the intervention design is relatively simple and does not cover more adversarial scenarios that may arise in real-world applications.

\section*{Acknowledgments}
We gratefully acknowledge the support of the Natural Sciences and Engineering Research Council of Canada (NSERC) under Grant ALLRP 585937-23 and Mitacs under Grant IT35587.

\bibliography{custom}

\appendix

\section{Data Statistics}
\label{sec:data}
This appendix section summarizes the dataset statistics used across our experiments. Table~\ref{tab:eventstoryline-stats} reports the five-fold split of the EventStoryLine dataset. Table~\ref{tab:ctb-maven-stats} presents the test-set distributions of two widely used benchmarks, Causal-TimeBank and MAVEN-ERE. Finally, Table~\ref{tab:inter-doc-stats} provides the statistics of the document-level (inter-sentence) subset of EventStoryLine.

\begin{table}[t]
\centering
\small
\setlength{\tabcolsep}{10pt}
\begin{tabular}{c|ccc}
\toprule
\multicolumn{4}{c}{\textbf{Training Set Statistics}} \\
\midrule
\textbf{Fold} & \textbf{Total} & \textbf{Pos} & \textbf{Neg} \\
\midrule
1 & 7021 & 2179 & 4842 \\
2 & 7491 & 2122 & 5369 \\
3 & 7378 & 2128 & 5250 \\
4 & 7629 & 2213 & 5416 \\
5 & 6697 & 2094 & 4603 \\
\midrule
\multicolumn{4}{c}{\textbf{Test Set Statistics}} \\
\midrule
\textbf{Fold} & \textbf{Total} & \textbf{Pos} & \textbf{Neg} \\
\midrule
1 & 4108 & 630  & 3478 \\
2 & 3244 & 648  & 2596 \\
3 & 3326 & 696  & 2630 \\
4 & 2404 & 528  & 1876 \\
5 & 4916 & 682  & 4234 \\
\bottomrule
\end{tabular}
\caption{Statistics of the five-fold split on EventStoryLine for both the training and test sets. ``Pos'' denotes causal samples, while ``Neg'' denotes non-causal samples.}
\label{tab:eventstoryline-stats}
\end{table}

\begin{table}[t]
\centering
\small
\setlength{\tabcolsep}{10pt}
\begin{tabular}{l|ccc}
\toprule
\textbf{Dataset} & \textbf{Total} & \textbf{Pos} & \textbf{Neg} \\
\midrule
Causal-TimeBank & 7656 & 298  & 7358 \\
MAVEN-ERE       & 19642 & 3359 & 16283 \\
\bottomrule
\end{tabular}
\caption{Statistics of the Causal-TimeBank and MAVEN-ERE datasets. ``Pos'' denotes causal event pairs, and ``Neg'' denotes non-causal event pairs.}
\label{tab:ctb-maven-stats}
\end{table}

\begin{table}[t]
\centering
\small
\setlength{\tabcolsep}{10pt}
\begin{tabular}{l|ccc}
\toprule
\textbf{Dataset} & \textbf{Total} & \textbf{Pos} & \textbf{Neg} \\
\midrule
EventStoryLine  & 13000  & 3000 & 10000 \\
\bottomrule
\end{tabular}
\caption{
Document-level (inter-sentence) subset of the EventStoryLine. ``Pos'' denotes causal event pairs, and ``Neg'' denotes non-causal event pairs.
}
\label{tab:inter-doc-stats}
\end{table}

\section{Training Hyperparameters}
\label{sec:training}
This appendix section provides the training details used in our experiments. Table~\ref{tab:training-hparams} summarizes the training hyperparameters and LoRA configuration used in our training process.

\begin{table}[t]
\centering
\small
\setlength{\tabcolsep}{10pt}
\begin{tabular}{l|c}
\toprule
\multicolumn{2}{c}{\textbf{General Training Hyperparameters}} \\
\midrule
Training Framework & TRL \\
Epochs & 1 \\
Per-device Batch Size & 1 \\
Gradient Accumulation Steps & 8 \\
Effective Batch Size & 8 \\
Learning Rate & $2\times10^{-4}$ \\
Scheduler & Cosine Annealing \\
FP16 Training & Yes \\
Gradient Checkpointing & Enabled \\
Packing (Sequence Packing) & Enabled \\
\midrule
\multicolumn{2}{c}{\textbf{LoRA Configuration}} \\
\midrule
LoRA Rank ($r$) & 8 \\
Scaling Factor ($\alpha$) & 16 \\
LoRA Dropout & 0.05 \\
\bottomrule
\end{tabular}
\caption{Training hyperparameters and LoRA configuration for fine-tuning.}
\label{tab:training-hparams}
\end{table}

\section{Visualization of Model Outputs}
\label{sec:vis}
This appendix section aims to qualitatively compare model outputs before and after fine-tuning.
Figure~\ref{fig:vis-1.5} presents an example for Qwen2.5-1.5B, while Figure~\ref{fig:vis-1} shows a corresponding example for LLaMA-3.2-1B, both with non-causal ground-truth labels. As shown in the following, the original models incorrectly infer a causal link, whereas the fine-tuned models correctly identify the absence of causality and provides a more faithful explanation.

\begin{figure*}[t]
\centering
\begin{tcolorbox}[
  colback=blue!10!white,
  colframe=gray!70,
  boxrule=0.8pt,
  arc=3pt,
  left=6pt,
  right=6pt,
  top=6pt,
  bottom=6pt,
  width=\textwidth
]
\textbf{Text:}  
\texttt{Mark Jay Heller told a judge that Lohan was settling in at Morningside Recovery and argued that the actress should be allowed to stay until a judge approves her placement.}

\vspace{0.6em}
\textbf{Question:}  
Is there a causal relationship between \texttt{<argued>} and \texttt{<approves>}?  
Ensuring your complete explanation comes first, then followed by the final answer in the strict format: [Final Answer: Yes/No].

\vspace{0.6em}
\textbf{Output before fine-tuning:}  
The passage indicates that Mark Jay Heller argued that Lohan should be allowed to stay at Morningside Recovery until a judge approves her placement. This implies that there is a causal relationship between the two actions: Heller's argument and the judge's approval.  
\textbf{[Final Answer: Yes]}

\vspace{0.6em}
\textbf{Output after fine-tuning:}  
The passage states that Mark Jay Heller argued that the actress should be allowed to stay until a judge approves her placement. While the argument expresses a request, it does not cause the judicial approval. The approval is a separate, future event not shown to be caused by the argument.  
\textbf{[Final Answer: No]}
\end{tcolorbox}

\caption{Qualitative comparison of Qwen2.5-1.5B outputs before and after fine-tuning on a non-causal event pair.}
\label{fig:vis-1.5}
\end{figure*}

\begin{figure*}[t]
\centering
\begin{tcolorbox}[
  colback=blue!10!white,
  colframe=gray!70,
  boxrule=0.8pt,
  arc=3pt,
  left=6pt,
  right=6pt,
  top=6pt,
  bottom=6pt,
  width=\textwidth
]
\textbf{Text:}  
\texttt{Mark Jay Heller told a judge that Lohan was settling in at Morningside Recovery and argued that the actress should be allowed to stay until a judge approves her placement.}

\vspace{0.6em}
\textbf{Question:}  
Is there a causal relationship between \texttt{<argued>} and \texttt{<approves>}?  
Ensuring your complete explanation comes first, then followed by the final answer in the strict format: [Final Answer: Yes/No].

\vspace{0.6em}
\textbf{Output before fine-tuning:}  
To answer the question accurately, we need to analyze the relationship between the words argued and approves. The word argued is a verb that means to express or state an opinion or point of view, often in a persuasive or argumentative manner. It implies a level of intensity or conviction in one's stance or position. On the other hand, approves is a verb that means to give consent, approval, or acceptance of something, often in a formal or official capacity. Given the context of the passage, Mark Jay Heller is arguing that Lohan should be allowed to stay at Morningside Recovery, which implies that he is advocating for her to be approved or permitted to stay. Therefore, there is a causal relationship between argued and approves, as the former is a means of expressing an opinion or point of view, and the latter is a result of that expression.
\textbf{[Final Answer: Yes]}

\vspace{0.6em}
\textbf{Output after fine-tuning:}  
The passage states that Mark Jay Heller ``argued that the actress should be allowed to stay until a judge approves her placement.'' Here, ``argued'' refers to Heller's action of presenting a case to the judge, while ``approves" describes a future action by the judge that is conditional on the stay. The phrase ``until a judge approves'' establishes a temporal condition (i.e., the stay is permitted only up to the point of approval), not a causal link. Heller's argument is a request for temporary permission pending the judge's independent decision. The approval is portrayed as a separate, not as a consequence of Heller's argument.
\textbf{[Final Answer: No]}
\end{tcolorbox}

\caption{Qualitative comparison of Llama3.2-1B outputs before and after fine-tuning on a non-causal event pair.}
\label{fig:vis-1}
\end{figure*}

\section{Additional Evaluation Results with Extended Metrics}
\label{sec:add_evaluation}
To provide a more comprehensive evaluation, we additionally report FPR, TNR, and MCC in this appendix. From Table~\ref{tab:extended-metrics}, we observe that fine-tuning with CoT traces generated by our pipeline leads to significantly lower FPR and higher TNR and MCC, indicating reduced causal hallucination and improved overall prediction quality.

\begin{table}[htbp]
\centering
\small
\setlength{\tabcolsep}{10pt}
\renewcommand{\arraystretch}{1.05}
\begin{tabular}{l|c}
\toprule
\multicolumn{2}{c}{\textbf{Qwen2.5-1.5B (Vanilla)}} \\
\midrule
FPR  & 88.80 \\
TNR & 11.20 \\
MCC  & 0.075 \\
\midrule
\multicolumn{2}{c}{\textbf{Qwen2.5-1.5B (Ours)}} \\
\midrule
FPR  & \textbf{37.13 ($\downarrow$)} \\
TNR  & \textbf{62.87 ($\uparrow$)} \\
MCC ( & \textbf{0.247 ($\uparrow$)} \\
\bottomrule
\end{tabular}
\caption{Additional evaluation results with FPR, TNR, and MCC on EventStoryLine. Lower FPR is better, while higher TNR and MCC are better. ``Vanilla'' refers to the original model and ``Our'' indicates fine-tuning using CoT trace data generated by our proposed pipeline.}
\label{tab:extended-metrics}
\end{table}

\end{document}